\documentclass[letterpaper]{article} 
\usepackage{aaai2026}  
\usepackage{times}  
\usepackage{helvet}  
\usepackage{courier}  
\usepackage[hyphens]{url}  
\usepackage{graphicx} 
\usepackage{natbib}  
\usepackage{caption} 
\usepackage{algorithm}
\usepackage{algpseudocode} 
\usepackage{graphicx}
\usepackage{multirow} 
\usepackage{booktabs, tabularx} 
\usepackage{amsfonts}

\usepackage{newfloat}
\usepackage{listings}

\usepackage{amsmath}

\DeclareCaptionStyle{ruled}{labelfont=normalfont,labelsep=colon,strut=off} 
\floatstyle{ruled}
\newfloat{listing}{tb}{lst}{}
\floatname{listing}{Listing}
\title{Beyond Adapter Retrieval: \\Latent Geometry-Preserving Composition via Sparse Task Projection}
\author{
    Pengfei Jin\textsuperscript{\rm 1}\footnote{First author. pjin1@mgh.harvard.edu},
    Peng Shu\textsuperscript{\rm 2}\footnote{Co-first author. peng.shu@uga.edu},
    Sifan Song\textsuperscript{\rm 1},
    Sekeun Kim\textsuperscript{\rm 1},
    Qing Xiao\textsuperscript{\rm 1},
    Cheng Chen\textsuperscript{\rm 3},
    Tianming Liu\textsuperscript{\rm 2},
    Xiang Li\textsuperscript{\rm 1},
    Quanzheng Li\textsuperscript{\rm 1}\footnote{Corresponding author. Li.Quanzheng@mgh.harvard.edu}
}
\affiliations{
    \textsuperscript{\rm 1}Center of Advanced Medical Computing and Analysis, Massachusetts General Hospital and Harvard Medical School\\ Boston, MA 02114, USA\\
    \textsuperscript{\rm 2}School of Computing, The University of Georgia\\ Athens, GA 30602, USA\\
    \textsuperscript{\rm 3}Department of Electrical and Electronic Engineering, The University of Hong Kong\\ Hong Kong, China

}

\usepackage{bibentry}
\nocopyright
\begin{document}

\maketitle

\begin{abstract}
Recent advances in parameter-efficient transfer learning have demonstrated the utility of composing LoRA adapters from libraries of pretrained modules. However, most existing approaches rely on simple retrieval heuristics or uniform averaging,  which overlook the latent structure of task relationships in representation space. We propose a new framework for adapter reuse that moves beyond retrieval, formulating adapter composition as a geometry-aware sparse reconstruction problem. Specifically, we represent each task by a latent prototype vector derived from the base model’s encoder and aim to approximate the target task prototype as a sparse linear combination of retrieved reference prototypes, under an $\ell_1$-regularized optimization objective. The resulting combination weights are then used to blend the corresponding LoRA adapters, yielding a composite adapter tailored to the target task. This formulation not only preserves the local geometric structure of the task representation manifold, but also promotes interpretability and efficient reuse by selecting a minimal set of relevant adapters. We demonstrate the effectiveness of our approach across multiple domains—including medical image segmentation, medical report generation and image synthesis. Our results highlight the benefit of coupling retrieval with latent geometry-aware optimization for improved zero-shot generalization.

\end{abstract}


\section{Introduction}
\label{sec:intro}

\begin{figure*}[t]
  \centering
  \includegraphics[width=0.95\linewidth]{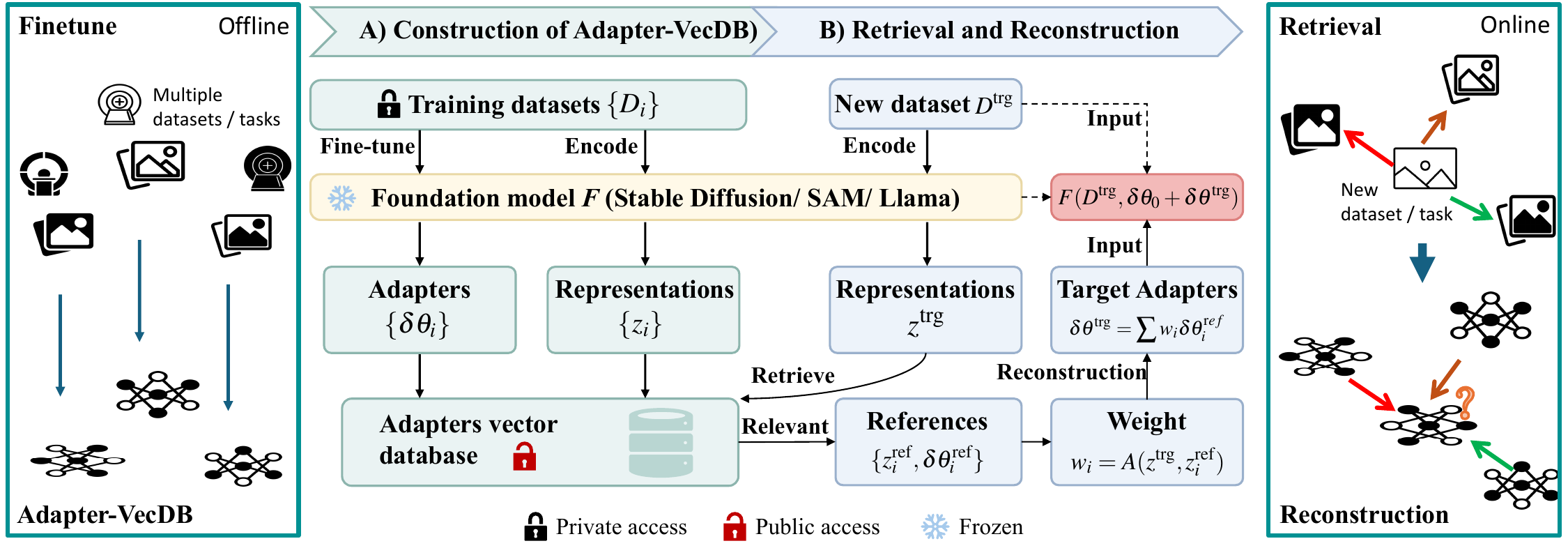}
  \caption{Overview of our latent geometry-aware adapter composition framework. Each task is represented by a prototype vector and stored in a vectorized adapter library (Adapter-VecDB), alongside its corresponding LoRA adapter. At inference time, we extract the target task prototype, retrieve the most relevant reference prototypes and adapters, and solve a sparse reconstruction problem in latent space. The resulting weights are then used to combine adapter parameters in a geometry-preserving manner.}
  \label{fig:pipeline}
\end{figure*}

Foundation models such as CLIP~\citep{radford2021learning}, LLaMA~\citep{touvron2023llama}, SAM~\citep{kirillov2023segment}, and Stable Diffusion~\citep{rombach2022high} have demonstrated impressive generalization across modalities and tasks. These models are pretrained on massive datasets and can serve as adaptable backbones for downstream applications in vision, language, and healthcare~ \citep{shu2024llms, zhao2024revolutionizing, rezayi2024exploring, yang2024examining}. However, adapting them to specific tasks remains costly in terms of data, compute, and retraining cycles.

Low-Rank Adaptation (LoRA)~\citep{hu2021lora} provides a scalable solution by fine-tuning a small set of task-specific parameters while freezing the base model. LoRA adapters can be cheaply trained, shared, and reused, enabling modular and community-driven adaptation. As open-source ecosystems like CivitAI and HuggingFace grow, large-scale adapter libraries have emerged—with over 100K public LoRAs reported for Stable Diffusion alone~\citep{luo2025stylus}. 

On the other hand, Retrieval-Augmented Generation (RAG) \citep{lewis2020retrieval,ma2023query,peng2024large,gao2022precise} enhances model outputs by integrating an external retrieval step, grounding responses in factual data. This method is especially effective in zero-shot learning scenarios, where the model encounters tasks it has not seen during training. This trend raises a natural question: \emph{Can we retrieve and reuse existing LoRA adapters to generalize to new tasks without additional training?}

Recent works have explored this idea from two angles. One line of research focuses on supervised or few-shot adapter fusion\citep{huang2023lorahub, prabhakar2024lora}, learning composition weights through gradient-based tuning or black-box optimization. While effective, these methods rely on task-specific labeled data, limiting their use in zero-shot settings. Another line centers on retrieval and reranking strategies\citep{zhao2024loraretriever, ostapenko2024towards}, selecting relevant adapters based on embedding similarity. However, these methods typically use uniform or heuristic similarity-based fusion, which fails to model the geometric structure of task relationships in latent space.

In this work, we shift the focus from retrieval to composition. We propose a latent geometry-aware framework that formulates adapter fusion as a sparse projection problem. Specifically, we represent each task by a prototype vector in representation space, and approximate the target prototype as a sparse linear combination of retrieved reference prototypes. The resulting weights—solved via $\ell_1$-regularized optimization—are used to combine the corresponding LoRA adapters in parameter space. This formulation captures the structure of the task manifold, promotes interpretable reuse, and requires no supervision.

Figure~\ref{fig:pipeline} illustrates our pipeline. We maintain a vectorized adapter library (Adapter-VecDB), which stores pretrained LoRA modules alongside their latent task representations. Given a new task, we extract its prototype, retrieve the $k$ most relevant adapters, and compute sparse combination weights via constrained reconstruction. Compared to averaging or similarity-based fusion, our method leverages the latent geometry of task space to guide composition.

We alidate our approach across three domains: medical image segmentation, radiology report generation, and image synthesis. Our method consistently outperforms baseline strategies under zero-shot evaluation, especially in distribution-shifted and low-resource settings.

Our main contributions are summarized as follows:
\begin{itemize}
    \item We propose a novel framework for geometry-aware adapter composition, based on sparse projection over task prototypes.
    \item Our formulation reveals that different fusion methods can be viewed as special cases under different assumptions about prototype geometry, offering a unified lens to understand adapter composition.
    \item We demonstrate strong zero-shot performance on multiple domains using only pretrained adapters without any additional data or fine-tuning.
\end{itemize}

\section{Preliminaries and Related Work}
\label{sec:RW}

\subsection{Retrieval and Zero-Shot Generalization}

Retrieval-augmented generation (RAG) methods enhance large models by injecting external information at inference time, typically through dynamic retrieval of text or structured data \citep{ma2023query,peng2024large,gao2022precise}. These approaches have improved low-resource performance and reduced hallucinations, especially in sensitive domains such as healthcare, where raw data access is limited \citep{seo2024retrieval,parvez2022retrieval}. However, RAG typically relies on textual corpora and does not support parameter-level reuse. Zero-shot learning, by contrast, aims to generalize to unseen tasks by leveraging semantic similarity or structural priors \citep{wang2019survey,xian2017zero,fu2018recent}. Prior works such as DeViSE~\citep{frome2013devise}, GCN-ZL~\citep{wang2018zero}, and DGP-ZL~\citep{kampffmeyer2019rethinking} learn mappings from task descriptions to model parameters via semantic or graph-based embeddings. Unlike these methods, we focus on composing pretrained adapters using latent task prototypes—without additional data or model training.

\subsection{Parameter Composition and Adapter Retrieval}

Modular parameter composition has emerged as a flexible alternative to full-model fine-tuning. Early methods such as AdapterFusion \citep{pfeiffer2021adapterfusion} learn supervised fusion weights to combine task-specific adapters, while AdapterSoup \citep{chronopoulou2023adaptersoup} performs zero-shot fusion by averaging retrieved adapters selected via domain similarity. Recent retrieval-based approaches aim to improve flexibility and generalization: LoRA-Hub \citep{huang2023lorahub} combines LoRA modules using gradient-free optimization with few-shot supervision; LoRA-Retriever \citep{zhao2024loraretriever} employs a dual-encoder model to retrieve and weight LoRA adapters for zero-shot transfer, supporting both parameter fusion and MoE-style output averaging. Ostapenko et al.\citep{ostapenko2024towards} route tokens to adapters via prototype alignment. Stylus\citep{luo2025stylus} retrieves adapters using prompt-VLM similarity and decomposes prompts into subtasks for fine-grained control. LoRA Soups~\citep{prabhakar2024lora} optimize adapter combination via learnable concatenation with few-shot supervision. 

Beyond retrieval-based composition, other works investigate parameter combination through expert gating and training-time routing. SMEAR \citep{muqeeth2023soft} and MoE-LoRA variants \citep{ma2024modula, fan2025make} use learned routers to combine adapter outputs dynamically at each layer. AdaMix \citep{wang2022adamix} applies stochastic routing and consistency regularization during training, while UP-RLHF \citep{zhai2023uncertainty} and IOP-FL \citep{jiang2023iop} optimize adapter selection through reinforcement or consistency-based objectives. Zoo-Tuning \citep{shu2021zoo} adapts pretrained model weights to target tasks using learned combination coefficients, and Model Soup \citep{wortsman2022model} explores simple averaging of models or adapters without retraining. LoRA-Ensemble \citep{halbheer2024lora} similarly averages adapters at inference time but focuses on distributional robustness. While these approaches vary in supervision and computational assumptions, most require either training or access to large-scale evaluation, in contrast to our plug-and-play, data-free composition strategy based on latent task geometry.

\subsection{Sparse Composition and Geometric Reconstruction}

Sparse composition techniques—such as compressed sensing~\citep{donoho2006compressed}, dictionary learning~\citep{mairal2010online}, and manifold learning~\citep{roweis2000nonlinear}—reconstruct inputs from a small number of basis elements while preserving local geometry. These ideas have inspired applications in multi-task learning~\citep{argyriou2008convex}, subspace clustering~\citep{elhamifar2013sparse}, and prototype-based adaptation in deep learning~\citep{zhao2023prototype,lee2019meta}, where task embeddings guide parameter reuse. However, few methods treat adapter composition explicitly as a geometry-aware sparse reconstruction problem. Our work fills this gap by formulating adapter fusion as an $\ell_1$-regularized projection over task prototypes, enabling zero-shot reuse without additional supervision.

\section{Method}
\label{sec:method}
We propose a three-stage retrieval-and-composition pipeline for zero-shot task adaptation. The key idea is to represent each task as a latent prototype vector in a shared subspace and to reconstruct the target task using a sparse combination of reference tasks. This latent-space reconstruction yields interpretable weights, which are then used to compose the corresponding adapters.

\begin{figure}[h]
  \centering
  \includegraphics[width=\linewidth]{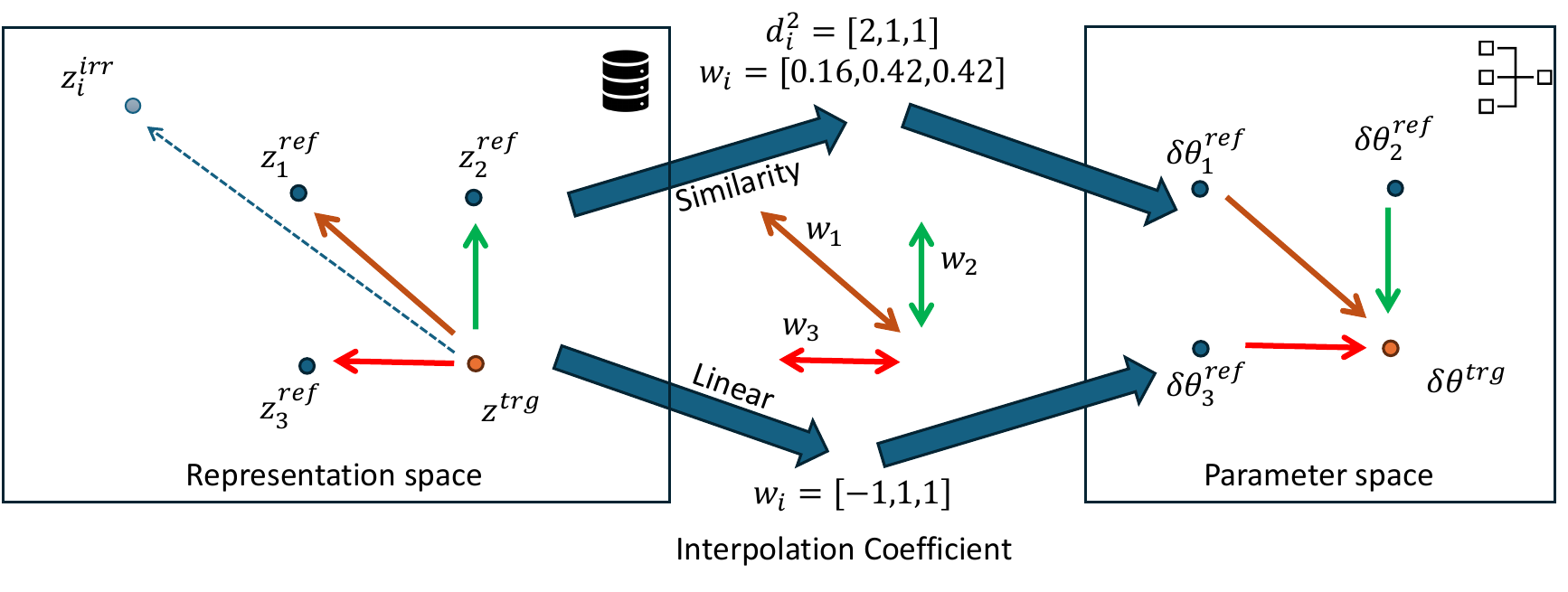}
  \caption{ Overview of our adapter composition framework. 
  (1) The input dataset is first encoded into a task prototype \( z^{\text{trg}} \). 
  (2) We retrieve a set of reference adapters and their prototypes \( \{z_i^{\text{ref}}, \delta \theta_i^{\text{ref}}\} \) from the Adapter-VecDB. 
  (3) Composition weights \( \{w_i\} \) are computed in the prototype space using one of different strategies: similarity-based softmax weighting or Least-Squares Linear Combination (4) The final adapter \( \delta \theta^{\text{trg}} \) is synthesized as a weighted sum in parameter space and applied to the frozen backbone.
  }
  \label{fig:workflow}
\end{figure}

\subsection{Task Prototype Construction}
\label{sec:method:prototype}

We begin by representing each task in a shared latent space via a fixed-length prototype vector. This prototype serves as a compact summary of the task’s distribution and forms the basis for downstream adapter composition. We organize these prototypes, along with their corresponding adapters, in a centralized database we refer to as the \textit{Adapter-VecDB}, analogous to the retrieval database in RAG systems. This database supports efficient lookup and geometric reasoning over previously trained adapters.

For a given dataset \( D_i \), we compute its task prototype \( z_i \in \mathbb{R}^d \) by averaging the feature representations of its constituent samples:
\begin{equation}
z_i = \frac{1}{|D_i|} \sum_{x_j \in D_i} E(x_j, \theta_0),
\label{eq:encoder}
\end{equation}
where \( E(\cdot, \theta_0) \) denotes a pretrained encoder (e.g., CLIP~\citep{radford2021learning}, LLaMA~\citep{touvron2023llama}) and \( \theta_0 \) is the frozen backbone model. This mean-pooling strategy ensures that the prototype preserves coarse semantic and statistical structure while remaining efficient to store and compute.

In practice, we adopt the encoder from the foundation model itself to ensure compatibility with the LoRA adapters. This approach aligns with the strategy used in the encoder component of the MoE, where feature maps serve a pivotal role in the model architecture. When full access to task datasets is unavailable—such as for public or third-party adapters like those from Stable Diffusion—we instead compute \( z_i \) from a small set of descriptive examples, metadata, or natural language summaries associated with the adapter. 

For simplicity and practicality in representing dataset features, we initially explored using various distribution distance metrics, such as the Chamfer distance \citep{borgefors1986distance}, Nearest Neighbor Distance \citep{alt1995computing}, Mean Distance \citep{carroll1998multidimensional}, to measure similarities between datasets. However, these metrics did not show significant differences in dataset characteristics. 

The collection \( \{z_i\} \) across available source tasks defines the \textit{prototype subspace}, which we treat as a geometric basis for reconstructing new, unseen tasks. This forms the foundation for our geometry-aware adapter composition described next. 

\begin{algorithm}[t]
\caption{Geometry-Aware Adapter Composition}
\label{alg}
\begin{algorithmic}[1]
\Require Foundation model \( F(\cdot, \theta_0) \); Adapter-VecDB \( \{ (z_i^{\text{ref}}, \delta \theta_i^{\text{ref}}) \}_{i=1}^N \); target dataset \( D^{\text{trg}} \)
\Ensure Adapted model \( F(\cdot, \theta^{\text{trg}}) \)

\State \( z^{\text{trg}} = \frac{1}{|D^{\text{trg}}|} \sum_{x_j \in D^{\text{trg}}} E(x_j, \theta_0) \) 
\Statex \phantom{s}\hfill $\triangleright$ Compute prototype for the target task

\State \( \{z_i^{\text{ref}}\}_{i=1}^k = \text{top-}k\text{-NN}(z^{\text{trg}}, \{z_j^{\text{ref}}\}) \)
\Statex \phantom{s} \hfill $\triangleright$ Retrieve $k$ nearest neighbors in the prototype space

\State \( \{w_i\} = \mathcal{A}(\{z_i^{\text{ref}}\}, z^{\text{trg}}) \)
\Statex \phantom{s} \hfill $\triangleright$ Compute composition weights (e.g., similarity, LS, sparse)

\State \( \delta \theta^{\text{trg}} = \sum_{i=1}^k w_i \delta \theta_i^{\text{ref}} \)
\Statex \phantom{s} \hfill $\triangleright$ Compose the adapter in parameter space

\State \( \theta^{\text{trg}} = \theta_0 + \delta \theta^{\text{trg}} \)
\Statex \phantom{s} \hfill $\triangleright$ Output the final adapted model

\end{algorithmic}
\end{algorithm}

\subsection{Geometry-Aware Sparse Composition}
\label{sec:method:composition}

Given a target task with prototype \( z^{\text{trg}} \), our goal is to compose a new adapter by selecting and combining relevant pretrained adapters from the Adapter-VecDB. We treat this as a geometry-aware reconstruction problem in the task prototype space, where the target prototype is approximated as a linear combination of retrieved source prototypes.

Formally, let \( \{z_i^{\text{ref}}\}_{i=1}^k \) be the prototypes of the top-$k$ retrieved adapters from the Adapter-VecDB, and let \( \{w_i\} \) denote the corresponding combination weights. We consider three strategies to compute the weights \( w_i \), each reflecting a different assumption about the structure of the latent space.

\paragraph{Similarity-Based Weighting.}
A common heuristic is to assume that similarity in the latent space correlates with transferability. We compute distances using the squared $\ell_2$ norm:
\begin{equation}
d^2(z_i, z^{\text{trg}}) = \|z_i - z^{\text{trg}}\|^2_2,
\label{eq:distance}
\end{equation}
and define weights via a softmax over the negative distances:
\begin{equation}
w_i = \frac{\exp(-\lambda_1 d_i^2)}{\sum_j \exp(-\lambda_1 d_j^2)},
\label{eq:sim_w}
\end{equation}
where \( \lambda_1 \) is a temperature parameter controlling the sharpness of the weighting distribution.

\paragraph{Least-Squares Linear Combination.}
Rather than relying on pairwise similarity, we can explicitly reconstruct the target prototype using a linear combination of the retrieved references. The weights are chosen to minimize reconstruction error under a sum-to-one constraint:
\begin{equation}
\min_{\sum w_i = 1} \left\| z^{\text{trg}} - \sum_{i=1}^k w_i z_i^{\text{ref}} \right\|_2^2.
\label{eq:lin_w}
\end{equation}
This strategy allows the model to interpolate among retrieved prototypes in a globally consistent way, but may still overfit when too many adapters are used.

\paragraph{Sparse Projection (Ours).}
To promote compact and interpretable compositions, we further impose an $\ell_1$ regularization term to encourage sparsity:
\begin{equation}
\min_{\sum w_i = 1} \left\| z^{\text{trg}} - \sum_{i=1}^k w_i z_i^{\text{ref}} \right\|_2^2 + \lambda_2 \| w \|_1,
\label{eq:linR_w}
\end{equation}
where \( \lambda_2 \) controls the trade-off between reconstruction fidelity and sparsity. This sparse formulation allows us to select a minimal subset of reference tasks that best explain the target, aligning with prior work in sparse coding and manifold reconstruction.

Different strategies yields a distinct set of weights \( \{w_i\} \) that guide the construction of the final adapter. The \textbf{uniform averaging} approach corresponds to prior work such as AdapterSoup~\citep{chronopoulou2023adaptersoup}, where selected adapters are merged using equal weights. The \textbf{similarity-based weighting} strategy resembles that of LoRA-Retriever~\citep{zhao2024loraretriever}, which assigns weights proportional to prototype similarity; while these works include additional contributions such as retrieval pipelines or routing mechanisms, we isolate and compare their implicit weighting strategies in our framework. In contrast to these heuristics, we propose a novel \textbf{sparse projection} method that formulates adapter composition as a constrained optimization problem in the latent space. 

Figure~\ref{fig:workflow} illustrates these strategies. While similarity-based methods assign positive weights based on proximity, linear combinations can capture more nuanced structure—including negative contributions—by reconstructing the target from the geometry of the prototype manifold. The experimental section further demonstrates how these differences impact generalization and performance.

\subsection{Adapter Reassembly}
\label{sec:method:reassembly}

Once the weights \( \{w_i\} \) are computed through geometry-aware composition in the prototype space, we apply them to merge the corresponding LoRA adapters in parameter space. Let \( \delta \theta_i^{\text{ref}} \) denote the LoRA weight update associated with the $i$-th retrieved adapter. The composite adapter for the target task is then given by:
\begin{equation}
\delta \theta^{\text{trg}} = \sum_{i=1}^k w_i \delta \theta_i^{\text{ref}}.
\label{eq:assemble}
\end{equation}
This aggregated update is added to the frozen base model \( \theta_0 \), yielding the final adapted model \( F(\cdot, \theta_0 + \delta \theta^{\text{trg}}) \).

Importantly, this process does not involve any gradient-based optimization or parameter tuning. The target adapter is generated entirely through linear combination of pretrained modules, guided by task geometry in representation space. Compared to training new adapters from scratch, our approach offers significant computational savings and eliminates the need for labeled data.

Moreover, the sparse nature of the weights enhances interpretability: only a small subset of reference adapters contributes to the final composition, allowing us to trace which source tasks influence the target. 

The complete pipeline is summarized in Algorithm~\ref{alg}, which outlines the construction of task prototypes, retrieval of candidate adapters, computation of geometry-aware weights, and reassembly of the composite adapter.

\section{Experiments}

\label{sec:Exp}

\subsection{Experimental Settings and Comparison Methods}
\label{sec:exp-settings}

\begin{table*}[h]
\centering
\begin{tabular}{ccc|cccc}
\toprule
\multirow{2}{*}{Dataset} & \multirow{2}{*}{Pre-trained} & \multirow{2}{*}{SFT} & \multicolumn{4}{c}{Zero-shot} \\
\cmidrule(lr){4-7}
 &  &  & Avg & Sim & Lin & Ours \\
 \midrule
Prostate-A~\citep{liu2020saml} & - & {95.4\%} & 80.3\% & 87.8\% & 86.3\% & \textbf{90.5\%} \\
Prostate-B~\citep{liu2020saml} & - & {92.8\%} & 77.5\% & 85.0\% & 83.4\% & \textbf{86.0\%} \\
Prostate-C~\citep{liu2020saml} & - & {90.5\%} & 51.0\% & 59.8\% & 61.9\% & \textbf{64.7\%} \\
Prostate-D~\citep{liu2020saml} & - & {91.2\%} & 74.9\% & 82.6\% & 86.7\% & \textbf{90.3\%} \\
Prostate-E~\citep{liu2020saml} & - & {92.7\%} & 64.6\% & 56.9\% & 52.0\% & \textbf{79.1\%} \\
Prostate-F~\citep{liu2020saml} & - & {93.0\%} & 82.2\% & 80.8\% & 82.4\% & \textbf{90.3\%} \\
AbdAtlas~\cite{li2024abdomenatlas} & - & {85.1\%} & 19.2\% & 22.9\% & 2.5\% & \textbf{64.5\%} \\
AutoPet~\cite{Gatidis2022}  & - & {88.9\%} & 52.7\% & 83.4\% & 85.9\% & \textbf{87.2\%} \\
Abdomen1k~\cite{ma2021abdomenct} & - & {88.6\%} & 36.6\% & 82.4\% & 80.5\% & \textbf{82.6\%} \\
RAOS~\cite{luo2024rethinking} & - & {88.5\%} & 58.6\% & 80.5\% & 78.6\% & \textbf{85.5\%} \\
\bottomrule
\end{tabular}
\caption{Comparison of DICE scores on the medical image segmentation task across different testing datasets.
All methods use the same $k$-NN retrieval pipeline; only the adapter combination strategy differs.
Avg: uniform averaging~\citep{chronopoulou2023adaptersoup};
Sim: softmax similarity-weighted combination~\citep{zhao2024loraretriever};
Lin: convex least-squares projection;
Ours: $\ell_1$-regularized sparse projection (ours).
The best performance in each zero-shot setting is highlighted in bold.
}
\label{tab:DICE}
\end{table*}

We evaluate our method across three domains—medical image segmentation, medical report generation, and image synthesis—using SAM~\citep{kirillov2023segment}, LLaMA 3.1 8B~\citep{dubey2024llama}, and Stable Diffusion v1.5~\citep{rombach2022high}. Each task poses a different challenge for adapter composition and zero-shot generalization.

For SAM and LLaMA, where no large-scale LoRA libraries exist, we simulate adapter reuse by training a small set of LoRAs on distinct medical datasets. Each dataset is treated as a task, with leave-one-out evaluation: LoRAs are trained on all but one dataset and composed to adapt to the held-out target. For Stable Diffusion, we follow Stylus~\citep{luo2025stylus} and use open LoRA library to build an Adapter-VecDB. At test time, we retrieve relevant adapters and compose them using different weighting strategies—without any fine-tuning—to generate images in a zero-shot fashion.

To ensure fair comparison, all methods share the same retrieval step ($k$-NN over prototypes), and only differ in how weights \( \{w_i\} \) are computed:
\begin{itemize}
    \item \textbf{Average (Avg):} Uniform averaging of all retrieved adapters, following AdapterSoup~\citep{chronopoulou2023adaptersoup}.
    \item \textbf{Similarity-weighted (Sim):} Softmax-weighted combination based on prototype similarity (Eq.~\eqref{eq:sim_w}), following Lora-Retriever~\citep{zhao2024loraretriever}, with $\lambda_1 = 1$.
    \item \textbf{Linear Combination (Lin):} Least-squares projection onto the subspace of retrieved prototypes (Eq.~\eqref{eq:lin_w}), with convex weight constraint ($\sum w_i = 1$) but no sparsity.
    \item \textbf{Sparse Projection (Ours):} $\ell_1$-regularized reconstruction (Eq.~\eqref{eq:linR_w}) that additionally encourages sparse adapter selection. We use $\lambda_2 = 10$ for vision tasks and $\lambda_2 = 1$ for language tasks, reflecting adapter availability.
\end{itemize}

All experiments are conducted using 8 NVIDIA H100 80GB GPUs, with adapter training performed under frozen backbones. Additional implementation details, dataset statistics, extended results, and visualizations are provided in the Appendix. Code and checkpoints will be released upon publication.

\subsection{Medical Image segmentation}

To evaluate our method in a high-variance domain, we consider prostate and abdominal CT segmentation using the SAM foundation model~\citep{kirillov2023segment}. We construct a collection of LoRA adapters, each trained on a distinct dataset from a different scanner, vendor, or population. This setting introduces significant domain shifts across datasets, making it a suitable testbed for zero-shot adaptation. We perform leave-one-out cross-validation: for each target dataset, we compose an adapter using those trained on all other datasets. The segmentation performance is measured using the Dice score~\citep{carass2020evaluating}. 

\begin{figure}[t]
  \centering
  \includegraphics[width=\linewidth]{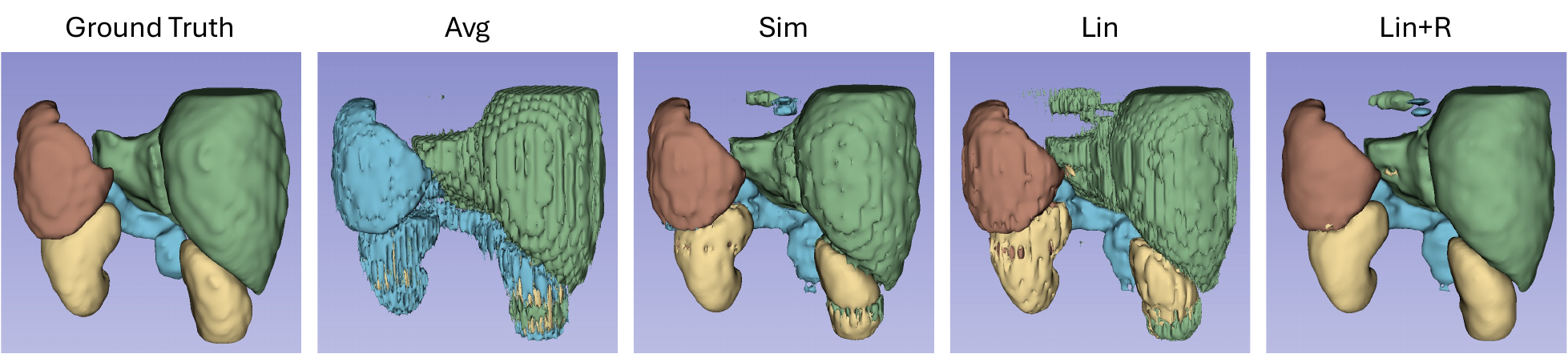}
  \caption{Visualization of segmentation results under different adapter fusion methods. Our method produces the most accurate and anatomically consistent segmentation, reducing both over- and under-segmentation artifacts.}
  \label{fig:seg_vis}
\end{figure}

\begin{table}[t]
\centering
\begin{tabular}{llllll}
\toprule
            & A     & B    & C    & D     & F    \\
\midrule
Sim   & 0.06  & 0.31 & 0.44 & 0.08  & 0.12 \\
Lin   & -1.13 & 0.67 & 0.26 & 0.11  & 1.09 \\
Ours  & -0.49 & 0.47 & 0.06 & -0.03 & 0.99\\
\bottomrule
\end{tabular}
\caption{Weight distribution of our methods applied to Prostate-E Segmentation, with columns representing reference Prostate datasets.}
\label{tab:Seg_E_w}
\end{table}

\paragraph{Quantitative results.}
Table~\ref{tab:DICE} reports the segmentation results for all methods. As expected, the pre-trained SAM model (without any adaptation) failed to produce meaningful results, highlighting the need for task-specific tuning. Supervised fine-tuning (SFT) serves as an upper bound but requires labeled data. Among zero-shot methods, our sparse projection approach consistently outperforms alternatives, achieving Dice scores close to SFT and significantly better than Avg or Sim baselines. The improvement is especially pronounced on out-of-distribution datasets. Visualization in Figure~\ref{fig:seg_vis} further confirm the anatomical accuracy and visual fidelity of our method.

\paragraph{Interpretability of composition.}
Table~\ref{tab:Seg_E_w} shows the weights assigned by different methods when adapting to the Prostate-E dataset. Similarity-based weights concentrate on Prostate-B/C, which are misleading in this case. The unregularized linear projection (Lin) yields unstable weights with large negative coefficients, indicating overfitting to noisy correlations. In contrast, our sparse projection selects a minimal and interpretable subset of relevant adapters, suppressing unreliable components and improving generalization. 

\begin{table*}[t]
\renewcommand{\arraystretch}{1.1} 
\setlength{\tabcolsep}{4pt} 
\centering
\begin{tabular}{llcc|cccc}
\toprule
\multirow{2}{*}{Modality
} & \multirow{2}{*}{Metric} &
\multirow{2}{*}{Pre-trained} & \multirow{2}{*}{SFT} &
\multicolumn{4}{c}{Zero-shot} \\ \cmidrule(lr){5-8}
 &  &  &  & Avg & Sim & Lin & Ours \\
\midrule
\multirow{2}{*}{CT Head} &
 ROUGE-L      & 0.201  & 0.2477 & 0.2124 & 0.2161 & 0.2140 & \textbf{0.2249} \\
 & BERTScore F1 & 0.8387 & 0.8499 & 0.8397 & 0.8412 & 0.8405 & \textbf{0.8433} \\
\midrule
\multirow{2}{*}{CT Abdomen} &
 ROUGE-L      & 0.1264 & 0.1387 & 0.1369 & 0.1374 & \textbf{0.1393} & 0.1379 \\
 & BERTScore F1 & 0.8039 & 0.8060 & 0.8068 & 0.8071 & 0.8076 & \textbf{0.8077} \\
\midrule
\multirow{2}{*}{MR} &
 ROUGE-L      & 0.1831 & 0.2153 & 0.1867 & 0.1914 & \textbf{0.1949} & 0.1896 \\
 & BERTScore F1 & 0.8365 & 0.8418 & 0.8378 & 0.8369 & 0.8380 & \textbf{0.8385} \\
\midrule
\multirow{2}{*}{X-ray} &
 ROUGE-L      & 0.1681 & 0.2159 & 0.1776 & 0.1794 & 0.1830 & \textbf{0.2084} \\
 & BERTScore F1 & 0.8494 & 0.8580 & 0.8521 & 0.8515 & 0.8522 & \textbf{0.8584} \\
\bottomrule
\end{tabular}
\caption{Performance comparison on the medical report impression generation task.
All methods share the same $k$-NN retrieval step over task prototypes; only the combination weights differ.
Avg: uniform averaging~\citep{chronopoulou2023adaptersoup};
Sim: softmax similarity-weighted combination~\citep{zhao2024loraretriever};
Lin: convex least-squares projection;
Ours: $\ell_1$-regularized sparse projection (ours).
    The best zero-shot result for each metric is shown in bold.}
\label{tab:all_results}
\end{table*}


\begin{table}[t]
\centering
\begin{tabular}{llll}
\toprule
          & CT (head) & MR   & XR   \\
\midrule
Sim & 0.34     & 0.33 & 0.33 \\
Lin & 0.80     & 0.18 & 0.02 \\
Ours & 0.82   & 0.18 & 0.00 \\
\bottomrule
\end{tabular}
\caption{Comparison of weight distributions in our similarity-based, linear combination and regularized linear combination methods for CT abdomen medical report impression task.}
\label{tab:CT_abdomen_w}
\end{table}

\subsection{Medical Report Impression}

We next evaluate our method in the domain of medical report summarization using Llama 3.1 8B as the foundation model~\citep{dubey2024llama}. Following~\citet{shi2024mgh}, we focus on the impression section of radiology reports across four datasets from Massachusetts General Hospital (MGH): CT-Head, CT-Abdomen, X-ray, and MR. Each dataset defines a task, and we conduct leave-one-out evaluation: for each test set, we compose adapters using the other three. We report ROUGE-L~\citep{lin2004rouge} and BERTScore~\citep{zhang2019bertscore} for lexical and semantic similarity.

\paragraph{Quantitative results.}
As shown in Table~\ref{tab:all_results}, our sparse projection (Ours) generally outperforms Avg, Sim, and Lin across most metrics and tasks. Notably, it also matches or even surpasses SFT in some cases, particularly when data distribution shift is moderate. The complete experiment results are shown in Appendix. While Lin occasionally attains the best ROUGE-L on a few tasks (e.g., CT-Abdomen and MR), our method delivers more balanced improvements overall. These improvements over Sim and Lin are especially meaningful given that all methods share the same retrieval pool, differing only in weight computation.

\paragraph{Interpretability of composition.}
To understand the behavior of each strategy, Table~\ref{tab:CT_abdomen_w} shows the adapter weights when generating impressions for CT-Abdomen. 
Our method assigns most weight to CT-Head (82\%) and modest weight to MR (18\%), consistent with clinical intuition. This validates the ability of sparse projection to select semantically aligned sources and avoid overfitting.

\subsection{{Image Generation}}
\begin{figure}[t]
  \centering
  \includegraphics[width=1.0\linewidth]{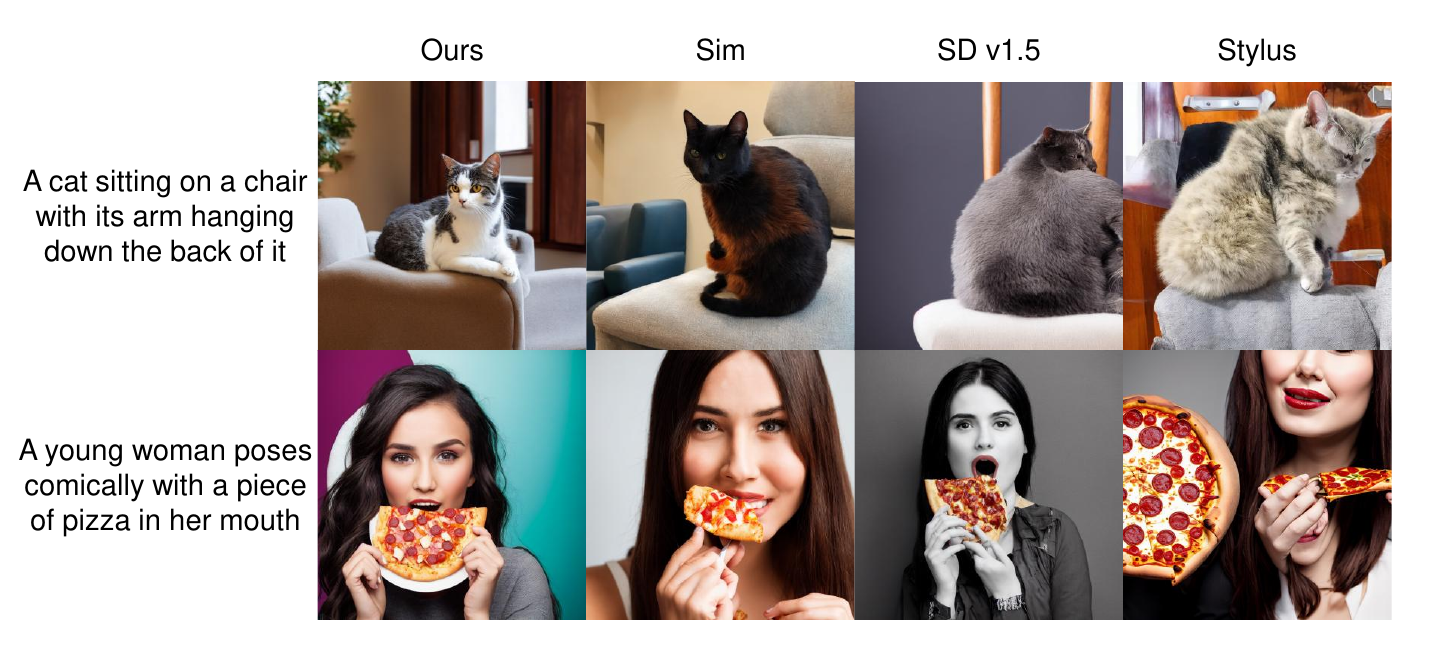}
  \caption{Examples of images generated by our method compared to Euclidean distance similarity, Stable Diffusion v1.5 baseline
and Stylus. Our ensemble model demonstrates improved adherence to input captions and produces higher-quality images.}
  \label{fig:SDpic}
\end{figure}

\begin{figure}[t]
  \centering
  \includegraphics[width=0.7\linewidth]{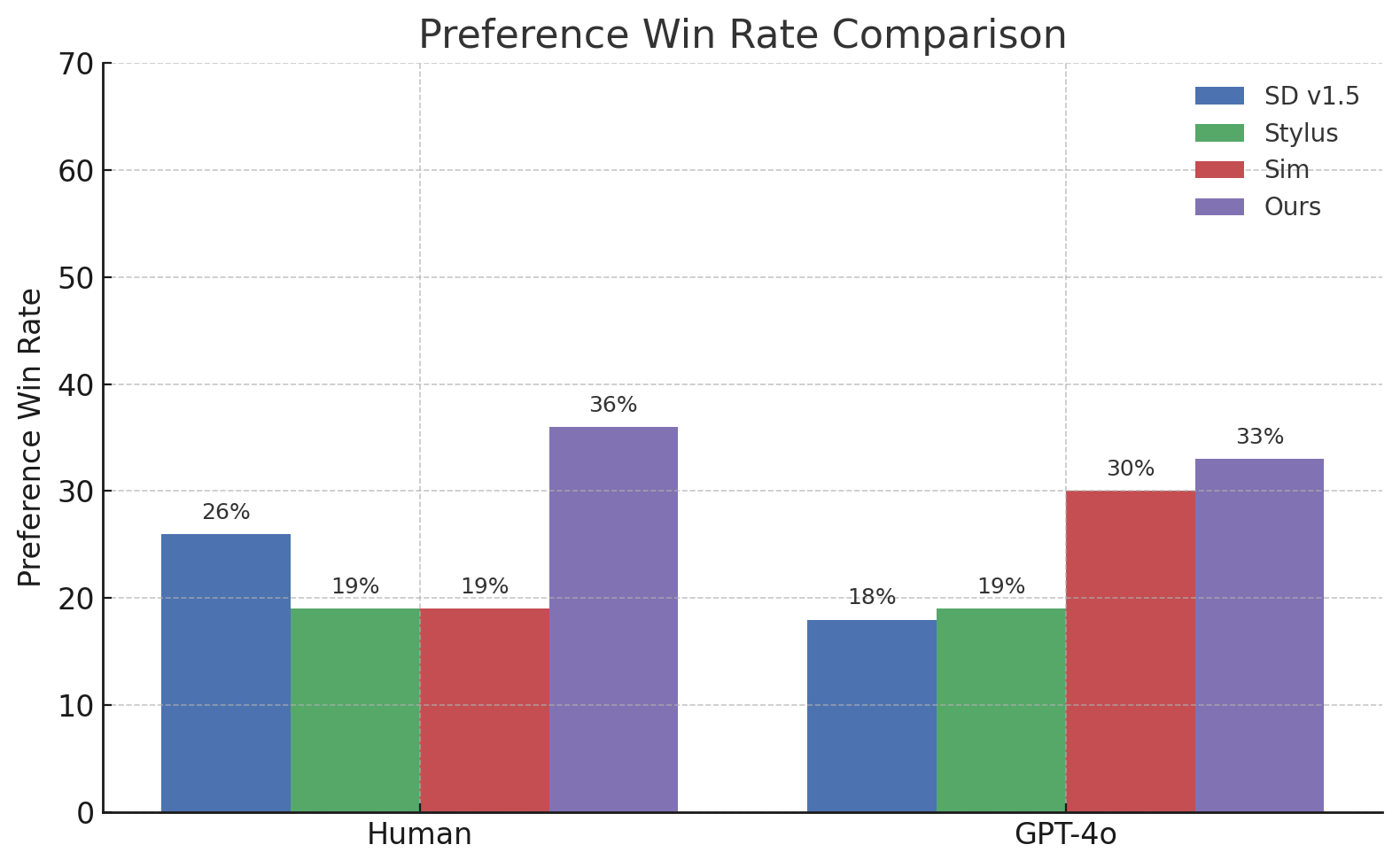}
  \caption{Human and AI evaluations. Our method obtains higher win rate on both tests.}
  \label{fig:SDeva}
\end{figure}

To assess our method in generative settings, we apply it to Stable Diffusion v1.5~\citep{rombach2022high}, leveraging the large-scale community adapter library from the Civitai platform, which is also employed by Stylus ~\citep{luo2025stylus}. Each adapter corresponds to a specific visual concept or style, and each prompt defines a new image generation task. Given a prompt and its paired reference image from the dataset, we retrieve relevant adapters via CLIP-encoded prototype similarity, where the image serves as auxiliary context to improve retrieval quality, and compose them using different strategies.

We compare four methods: the base model (without LoRA), Stylus~\citep{luo2025stylus} (retrieval + CLIP reranking), similarity-weighted (Sim), and our sparse projection (Ours). For evaluation, we follow Stylus and sample 100 prompts from MS COCO~\citep{lin2014microsoft}. Each method generates one image per prompt, and we conduct both GPT‑4o and human preference studies. In each evaluation trial, annotators (or GPT) are shown four images side‑by‑side—one from each method. Then, they are asked to select the best image based on alignment with the prompt as well as overall visual quality, including the absence of disfigured limbs and unrealistic object compositions.

Figure~\ref{fig:SDeva} shows that both GPT‑4o and human raters consistently prefer our method over Stylus, Sim, and the base model. Qualitative examples in Figure~\ref{fig:SDpic} demonstrate that our method produces images that are more semantically consistent with the prompts and exhibit noticeably higher visual quality, with fewer disfigured limbs and unrealistic object compositions. These improvements are achieved even when building upon the pruned SD v1.5 backbone, highlighting the effectiveness of our adapter composition strategy. See Appendix for details on the setup and additional examples.

\subsection{Ablation Study}
We conduct ablation experiments to better understand the behavior and flexibility of our proposed sparse adapter composition framework. Specifically, we examine (1) whether simply using the most similar adapter is sufficient, (2) whether composition can improve supervised LoRA when in-domain adapters are available, and (3) how our method compares in terms of training cost.

\begin{table}[t]
\centering
\resizebox{\columnwidth}{!}{
\begin{tabular}{lllllll}
\toprule & A     & B    & C    & D  & E   & F    \\
\midrule
DICE   & 90.5\%  & 86.4\% & 54.6\% & 90.0\%& 0.1\%  & 91.0\% \\
\bottomrule
\end{tabular}
}
\caption{Performance of Prostate dataset only the single nearest LoRA. Results are inconsistent across tasks.}
\label{tab:Top_LoRA}
\end{table}
\paragraph{Nearest adapter vs. composition.}
A natural baseline for zero-shot transfer is to retrieve only the most similar adapter (i.e., $k{=}1$) and directly apply it to the new task. Table~\ref{tab:Top_LoRA} reports the DICE scores when using only the nearest adapter for each test set. While this works reasonably well for some tasks (e.g., Prostate-A, D, F), it fails completely for others (e.g., Prostate-E), likely due to misleading similarity caused by distribution shifts. In contrast, our sparse ensemble method is more robust, as it integrates multiple references while suppressing irrelevant ones. This highlights the importance of geometry-aware composition over naïve nearest-neighbor selection.

\begin{table}[t]
\centering
\begin{tabular}{lllllll}
\toprule
            & A     & B    & C    & D  & E   & F    \\
\midrule
weight   & -0.21  & -0.07 & 1.10 & 0.05 & 0.03 & 0.11  \\
\bottomrule
\end{tabular}
\caption{Linear combination weights for Prostate-C, including its own supervised LoRA. The method selectively blends complementary knowledge.}
\label{tab:weight with self}
\end{table}

\paragraph{Enhancing supervised LoRA via cross-task composition.}
Although our method is designed for zero-shot scenarios where no in-domain LoRA is available, it can also enhance performance when a supervised LoRA exists. By including the supervised adapter in the candidate pool during composition, we evaluate whether cross-task fusion offers benefits. As shown in Table~\ref{tab:weight with self}, our method assigns the highest weight to the in-domain LoRA (C) while also incorporating support from others (F), and assigning negative weights to less relevant ones (A). This yields a slight improvement in Dice score (90.8\% vs. 90.5\%), illustrating that our approach can flexibly refine existing adapters by leveraging complementary knowledge—without requiring additional retraining.

\subsubsection{Efficiency of retrieval-based composition.}
Finally, we compare the training cost between our method and conventional supervised fine-tuning. For example, training a LoRA on medical report data takes approximately 30 minutes on 8 NVIDIA H100 GPUs. In contrast, retrieving relevant adapters and solving our sparse projection takes under 3 minutes per task—less than 10\% of the time. For large-scale systems or privacy-sensitive domains, this efficiency makes retrieval-based composition a practical and scalable alternative to fine-tuning.

\section{Discussion and Conclusion}
\label{sec:discussion}

We introduce a geometry-aware framework for adapter composition by modeling adapter reuse as a sparse reconstruction problem in latent task space. This approach enables interpretable, plug-and-play zero-shot transfer and achieves strong results across multiple domains.

Unlike prior methods based on averaging or similarity-weighted fusion, our sparse projection strategy offers finer control over adapter selection and greater robustness in out-of-distribution settings. Ablation studies show our method can also enhance supervised adapters without retraining.

However, the effectiveness of our approach depends on the diversity and coverage of Adapter-VecDB, which in turn relies on contributions from the broader open-source community. In domains with limited adapters, this remains a constraint. Scaling to larger libraries will also require faster retrieval and more efficient optimization, such as through sparse indexing or prototype clustering. Future directions include jointly learning task prototypes, improving scalability, and extending to multi-modal settings. As adapter ecosystems grow, geometry-aware composition offers a practical and privacy-preserving alternative to traditional fine-tuning.

\section*{Acknowledgments}
Quanzheng Li’s research is supported in part by the National Institutes of Health under Grant
R01HL159183.

\bibliography{aaai2026}

\clearpage
\appendix
\setcounter{page}{1}

\section*{Appendix}
\addcontentsline{toc}{section}{Appendix}

\section{Construction of Adapter-VecDB}
\label{sec:app_vecDB}
For SAM model, we focus on medical image segmentation tasks. Consistent with the MA-SAM framework \citep{chen2023ma}, we use the same hyperparameter settings. We reproduce and train six prostate MRI adapters and four multi-organ abdomen CT adapters from different datasets~\citep{liu2020saml,ma2021abdomenct,Gatidis2022,luo2024rethinking,li2024abdomenatlas}. 
For both tasks, each dataset is iteratively treated as the target dataset, while the remaining datasets serve as reference datasets for zero-shot learning.

For Llama 3.1 8B model, we evaluate its performance on generating medical report impressions from provided findings. Specifically, we fine tune four LoRA models derived from the pre-trained Llama 3.1 8B model using four distinct datasets collected from Massachusetts General Hospital (MGH). These datasets comprise 24,801 CT abdomen reports, 63,745 CT head reports, 18,157 MR image reports, and 60,000 X-ray image reports. Each report includes detailed image findings and corresponding impressions. We create 20 different instructions asking for impressions and remove all the names in the reports by using regular expression. The fine-tuning process employ consistent hyperparameter settings: training batch size = 8, gradient accumulation steps = 4, optimizer = paged adamw 32bit, learning rate = $5*10^{-6}$, weight decay = 0.001, maximum gradient normal = 0.3, LoRA r = 16, LoRA alpha = 0.05. The number of training epochs is set as follows: 2 for CT abdomen, 1 for CT head, 3 for MR, and 1 for X-ray reports. In testing, we collecte 200 new reports for each type of medical image.

For StableDiffusion-v1.5 model, we build upon the work of \citep{luo2025stylus} by incorporating adapters and model checkpoints sourced from Civitai. As the ecosystem continues to expand, the number of available adapters has surpassed 100,000, with LoRA emerging as the predominant fine-tuning approach. To construct our LoRA database, we randomly sample 10,000 adapters from Civitai, ensuring a diverse representation of the available fine-tuning strategies.

\section{More Experiments and Visualization}
\subsection{Medical Image segmentation}

\begin{table*}[h]
\centering
\begin{tabular}{lllllll}
\toprule
 Dataset & Institution & Case & strength(T) & Resolution (mm) & Endorectal Coil & Manufactor  \\
\midrule

Prostate-A  & RUNMC       & 30             & 3           & 0.6-0.625/3.6-4                      & Surface                                                       & Siemens    \\
Prostate-B  &  BMC & 30             & 1.5         & 0.4/3                                & Endorectal                                                    & Philips    \\
Prostate-C  & HCRUDB      & 19             & 3           & 0.67-0.79/1.25                       & No                                                            & Siemens    \\
Prostate-D  & UCL         & 13             & 1.5 and 3   & 0.325-0.625/3-3.6                    & No                                                            & Siemens    \\
Prostate-E  & BIDMC       & 12             & 3           & 0.25/2.2-3                           & Endorectal                                                    & GE         \\
Prostate-F  & HK          & 12             & 1.5         & 0.625/3.6                            & Endorectal & Siemens \\
\bottomrule
\end{tabular}
\caption{Characteristics of Prostate MRI datasets from multiple institutions used in the study. This table details variations in magnetic field strength, spatial resolution, usage of endorectal coils, and MRI equipment manufacturers across six different sites, highlighting the diversity of data sources in our experiments.}
\label{tab:data_S}
\end{table*}

Table \ref{tab:data_S} illustrates the variability among different data sources used in our prostate experiments. The Prostate MR datasets differ significantly in terms of strength, resolution, and manufacturer, leading to notable shifts in data distribution. Using a single LoRA for segmentation tasks tends to result in overfitting to specific data distributions and fails to generalize across diverse datasets.

To quantify the impact of these distribution shifts, we analyzed the Euclidean distances $||z_i-z_j||^2_2$ between different training and testing sets, as detailed in Table \ref{tab:dis_D}. Each row in this table shows how one testing set differs from other training sets. Correspondingly, Table \ref{tab:DICE_D} displays the DICE scores achieved when applying models trained on these various sets to a given testing set. The results highlight the challenges posed by dataset variability and underscore the necessity for adaptive segmentation strategies that can effectively handle diverse data characteristics.

\begin{table*}[h]
\centering
\begin{tabular}{lllllll}
\toprule
  & Prostate-A        & Prostate-B       & Prostate-C       & Prostate-D        & Prostate-E       & Prostate-F       \\
\midrule
Prostate-A & \textbf{85.0758} & 94.8546 & 95.2915 & 89.6767  & 95.514  & 87.1439 \\
Prostate-B & 97.6358  & \textbf{89.4471} & 96.5984 & 95.2394  & 90.5619 & 98.885  \\
Prostate-C & 97.2556  & 97.017  & \textbf{85.7178} & 97.4879  & 95.1096 & 98.1607 \\
Prostate-D & \textbf{90.9688}  & 94.3976 & 96.8321 & 92.7221  & 94.3723 & 92.5209 \\
Prostate-E & 101.5153 & 96.0675 & 94.905  & 100.7156 & \textbf{82.0723} & 99.3386 \\
Prostate-F & 87.463   & 93.2918 & 95.6369 & 91.5755  & 95.074  & \textbf{86.0333} \\
\bottomrule
\end{tabular}
\caption{Euclidean distances between feature vectors of different Prostate datasets, quantifying the distribution shifts. Each entry represents the squared Euclidean distance $||z_i - z_j||^2_2$ between testing sets and trainig sets across sites A through F. The closest distances are highlighted in bold.}
\label{tab:dis_D}
\end{table*}

\begin{table*}[h]
\centering
\begin{tabular}{lllllll}
\toprule
  & Prostate-A        & Prostate-B       & Prostate-C       & Prostate-D        & Prostate-E       & Prostate-F       \\
\midrule
Prostate-A & \textbf{95.4}\% & 92.4\% & 44.3\% & 91.0\% & 83.3\% & 90.5\% \\
Prostate-B & 84.1\% & \textbf{92.8}\% & 44.8\% & 87.0\% & 86.4\% & 85.3\% \\
Prostate-C & 26.1\% & 60.2\% & \textbf{90.5}\% & 75.1\% & 54.6\% & 39.0\% \\
Prostate-D & 90.0\% & 86.7\% & 49.9\% & \textbf{91.2}\% & 71.5\% & 76.4\% \\
Prostate-E & 75.5\% & 84.8\% & 0.1\%  & 76.8\% & \textbf{92.7}\% & 85.8\% \\
Prostate-F & 91.0\% & 87.4\% & 58.2\% & 84.3\% & 90.1\% & \textbf{93.0}\% \\
\bottomrule
\end{tabular}
\caption{DICE scores for models tested across different Prostate datasets, reflecting model performance variability. The highest scores are highlighted in bold.}
\label{tab:DICE_D}
\end{table*}

\begin{figure*}[t]
  \centering
  \includegraphics[width=0.5\linewidth]{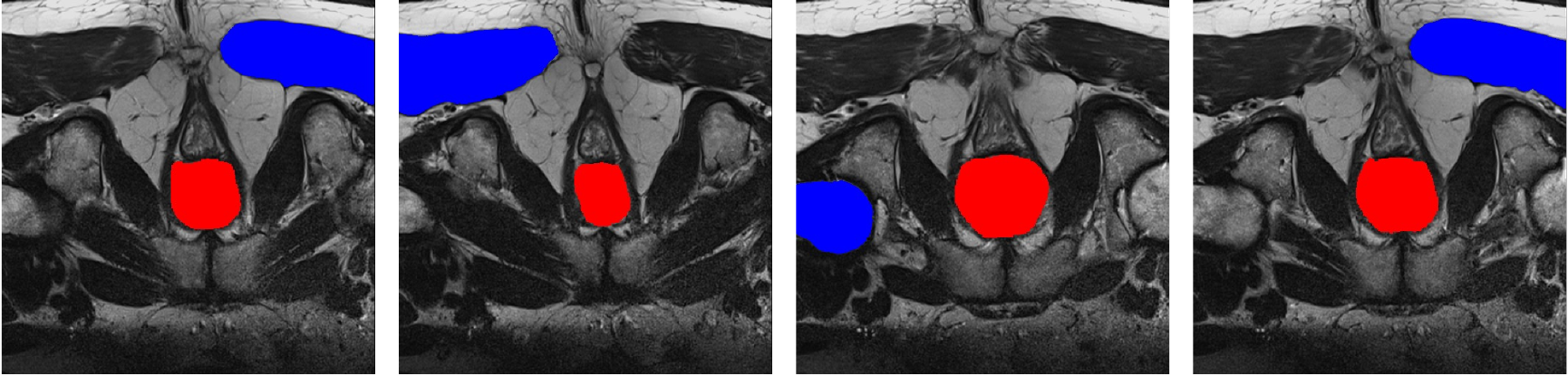}
  \caption{Pre-trained SAM segmentation outputs without the use of LoRA. The blue regions represent the segmentation results produced by SAM, while the red regions indicate the ground truth labels. This figure illustrates the randomness in organ selection by SAM when it lacks LoRA's task-specific guidance, highlighting the necessity of employing LoRA to ensure accurate and consistent organ segmentation across varying datasets.}
  \label{fig:SAM_noL}
\end{figure*}

The implementation of the SAM without LoRA was found to be ineffective, as SAM lacked the necessary guidance on which organs should be segmented. As illustrated in the examples shown in Figure \ref{fig:SAM_noL}, the organs targeted by SAM for segmentation appeared to be selected randomly. In contrast, LoRAs inherently contain task-specific information, such as the identification of the organs that need to be segmented.

Despite the presence of distribution shifts across different datasets, the organ categories required for segmentation remain consistent. This consistency is crucial, as it underlines why employing LoRA enables the completion of tasks that pre-trained models without retrieval capabilities fail to achieve. This finding demonstrates the importance of integrating task-specific knowledge in the form of LoRAs to guide the segmentation process effectively, particularly when dealing with diverse medical imaging datasets.

\begin{figure*}[t]
  \centering
  \includegraphics[width=\linewidth]{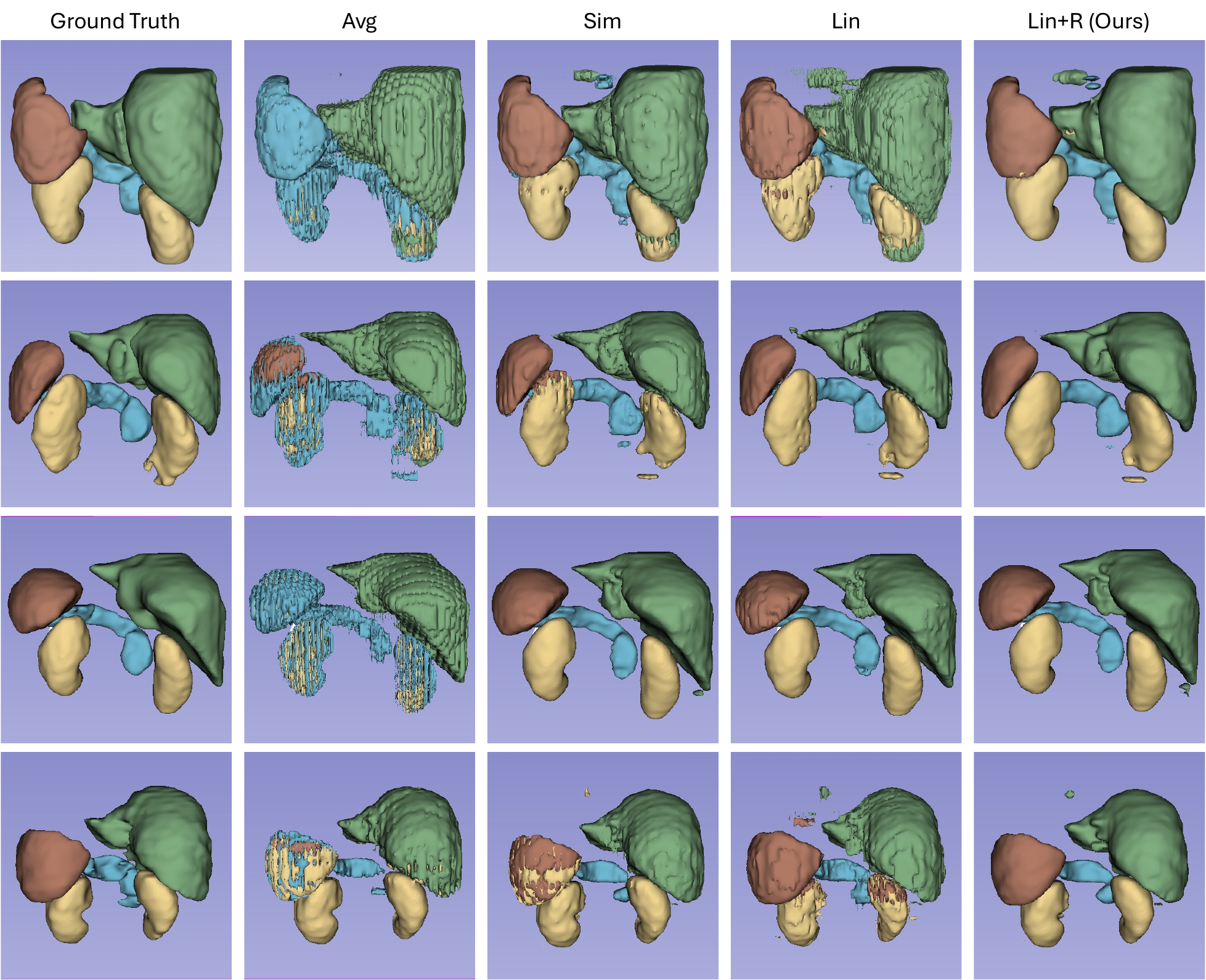}
  \caption{More Visualization of segmentation results under different adapter fusion methods. Our method produces the most accurate and anatomically consistent segmentation, reducing both over- and under-segmentation artifacts.}
  \label{fig:seg_vis_more}
\end{figure*}

To further illustrate the qualitative differences across fusion methods, we present additional segmentation visualizations in Figure~\ref{fig:seg_vis_more}. Compared to baselines, our method consistently produces anatomically plausible results, minimizing both over-segmentation (e.g., hallucinated regions) and under-segmentation (e.g., missing structures), especially under domain shifts. These examples highlight the benefits of sparse geometry-aware composition in ensuring robust and accurate model behavior across heterogeneous datasets.

\subsection{Medical Report Generation}
\label{other_llms}
Table \ref{tab:CT_head} to Table \ref{tab:XR_w} shows the experiment results in other medical image and corresponding weight in our methods. Our results suggest that the SFT model serves as an approximate upper bound for performance in these experiments. Both of our proposed methods consistently outperform the zero-shot pre-trained Llama 3.1 8B model, and in several cases achieve performance comparable to, or even surpassing, the SFT model, highlighting the effectiveness of our designs.

\begin{table*}[t]
\centering
\begin{tabularx}{\textwidth}{p{3cm} X X| X X X X}
\toprule
\multirow{2}{*}{Metrics} & \multirow{2}{*}{Pre-trained} & \multirow{2}{*}{SFT} & \multicolumn{4}{c}{Zero-shot} \\
\cmidrule(lr){4-7}
 &  &  & Avg & Sim & Lin & Ours \\
\midrule
ROUGE-L & 0.201 & 0.2477 & 0.2124 & 0.2161 & 0.2140 & \textbf{0.2249} \\
BertScore Precision & 0.8166 & 0.8278 & 0.8194 & 0.8202 & 0.8201 & \textbf{0.8219} \\
BertScore Recall & 0.8625 & 0.8739 & 0.8617 & 0.8640 & 0.8629 & \textbf{0.8666}\\
BertScore F1 & 0.8387 & 0.8499 & 0.8397 & 0.8412 & 0.8405 & \textbf{0.8433}\\
\bottomrule
\end{tabularx}
\caption{Performance comparison of our models against pre-trained Llama 3.1 8B, LoRA Supervised Fine-tuning (SFT), and zero-shot models on CT head medical report impression task.}
\label{tab:CT_head}
\end{table*}

\begin{table*}[t]
\centering
\begin{tabular}{llll}
\toprule
          & CT (abdomen) & MR   & XR   \\
\midrule
Sim & 0.32     & 0.32 & 0.36 \\
Lin & 0.25     & 0.33 & 0.42 \\
Ours & 0.71     & 0.29 & 0.00 \\
\bottomrule
\end{tabular}
\caption{Comparison of weight distributions in our similarity-based, linear combination and regularized linear combination methods for CT head medical report impression task.}
\label{tab:CT_head_w}
\end{table*}

\begin{table*}[t]
\centering
\begin{tabularx}{\textwidth}{p{3cm} X X| X X X X}
\toprule
\multirow{2}{*}{Metrics} & \multirow{2}{*}{Pre-trained} & \multirow{2}{*}{SFT} & \multicolumn{4}{c}{Zero-shot} \\
\cmidrule(lr){4-7}
 &  &  & Avg & Sim & Lin & Ours \\
\midrule
ROUGE-L & 0.1264 & 0.1387 & 0.1369 & 0.1374 & \textbf{0.1393} & 0.1379 \\
BertScore Precision & 0.7779 & 0.7789 & 0.7811 & 0.7815 & 0.7816 & \textbf{0.7819} \\
BertScore Recall & 0.8321 & 0.8355 & 0.8348 & 0.8350 & \textbf{0.8358} & \textbf{0.8358}\\
BertScore F1 & 0.8039 & 0.806 & 0.8068 & 0.8071 & 0.8076 & \textbf{0.8077} \\
\bottomrule
\end{tabularx}
\caption{Performance comparison of our models against pre-trained Llama 3.1 8B, LoRA Supervised Fine-tuning (SFT), and zero-shot models on CT abdomen medical report impression task.
}
\label{tab:CT_abdomen}
\end{table*}

\begin{table*}[t]
\centering
\begin{tabular}{llll}
\toprule
          & CT (head) & MR   & XR   \\
\midrule
Sim & 0.34     & 0.33 & 0.33 \\
Lin & 0.80     & 0.18 & 0.02 \\
Ours & 0.82   & 0.18 & 0.00 \\
\bottomrule
\end{tabular}
\caption{Comparison of weight distributions in our similarity-based, linear combination and regularized linear combination methods for CT abdomen medical report impression task.}
\label{tab:CT_abdomen_w1}
\end{table*}

\begin{table*}[t]
\centering
\begin{tabularx}{\textwidth}{p{3cm} X X |X X X X}
\toprule
\multirow{2}{*}{Metrics} & \multirow{2}{*}{Pre-trained} & \multirow{2}{*}{SFT} & \multicolumn{4}{c}{Zero-shot} \\
\cmidrule(lr){4-7}
 &  &  & Avg & Sim& Lin & Ours \\
\midrule
ROUGE-L & 0.1831 & 0.2153 & 0.1867 & 0.1914 & \textbf{0.1949} & 0.1896\\
BertScore Precision & 0.8107 & 0.8186 & \textbf{0.8128} & 0.8109 & 0.811  & 0.8118\\
BertScore Recall & 0.8644 & 0.8669 & 0.8649 & 0.8651 & 0.8671 & \textbf{0.8674}\\
BertScore F1 & 0.8365 & 0.8418 & 0.8378 & 0.8369 & 0.8380 & \textbf{0.8385} \\
\bottomrule
\end{tabularx}
\caption{Performance comparison of our models against pre-trained Llama 3.1 8B, LoRA Supervised Fine-tuning (SFT), and zero-shot models on MR medical report impression task.}
\label{tab:MR}
\end{table*}

\begin{table*}[t]
\centering
\begin{tabular}{llll}
\toprule
          & CT (abdomen) & CT (head)   & XR   \\
\midrule
Sim & 0.33    & 0.34 & 0.33 \\
Lin & 0.79     & 0.17 & 0.04 \\
Ours & 0.27     & 0.72 & 0.00 \\
\bottomrule
\end{tabular}
\caption{Comparison of weight distributions in our similarity-based, linear combination and regularized linear combination methods for MR medical report impression task.}
\label{tab:MR_w}
\end{table*}

\begin{table*}[t]
\centering
\begin{tabularx}{\textwidth}{p{3cm} X X |X X X X}
\toprule
\multirow{2}{*}{Metrics} & \multirow{2}{*}{Pre-trained} & \multirow{2}{*}{SFT} & \multicolumn{4}{c}{Zero-shot} \\
\cmidrule(lr){4-7}
 &  &  & Avg &Sim & Lin & Ours \\
\midrule
ROUGE-L & 0.1681 & 0.2159 & 0.1776 & 0.1794 & 0.1830  & \textbf{0.2084}  \\
BertScore Precision & 0.8244 & 0.837 & 0.8290 & 0.8273 & 0.8289 & \textbf{0.8408} \\
BertScore Recall & 0.8765 & 0.8807 & 0.877 & \textbf{0.8778} & 0.8774  & 0.8775\\
BertScore F1 & 0.8494 & 0.858 & 0.8521 & 0.8515 & 0.8522 & \textbf{0.8584}  \\
\bottomrule
\end{tabularx}
\caption{Performance comparison of our models against pre-trained Llama 3.1 8B, LoRA Supervised Fine-tuning (SFT), and zero-shot models on X-ray medical report impression task.}
\label{tab:XR}
\end{table*}

\textbf{\begin{table*}[t]
\centering
\begin{tabular}{llll}
\toprule
          & CT (abdomen) & CT (head)   & MR   \\
\midrule
Sim & 0.32     & 0.37 & 0.31 \\
Lin & 0.48     & 0.15 & 0.38 \\
Ours & 0.00     & 0.01 & 0.99 \\
\bottomrule
\end{tabular}
\caption{Comparison of weight distributions in our similarity-based, linear combination and regularized linear combination methods for X-ray medical report impression task.}
\label{tab:XR_w}
\end{table*}}

\subsection{Image Synthesis}
\label{Generation Results}

\begin{figure*}[h]
  \centering
  \includegraphics[width=\linewidth]{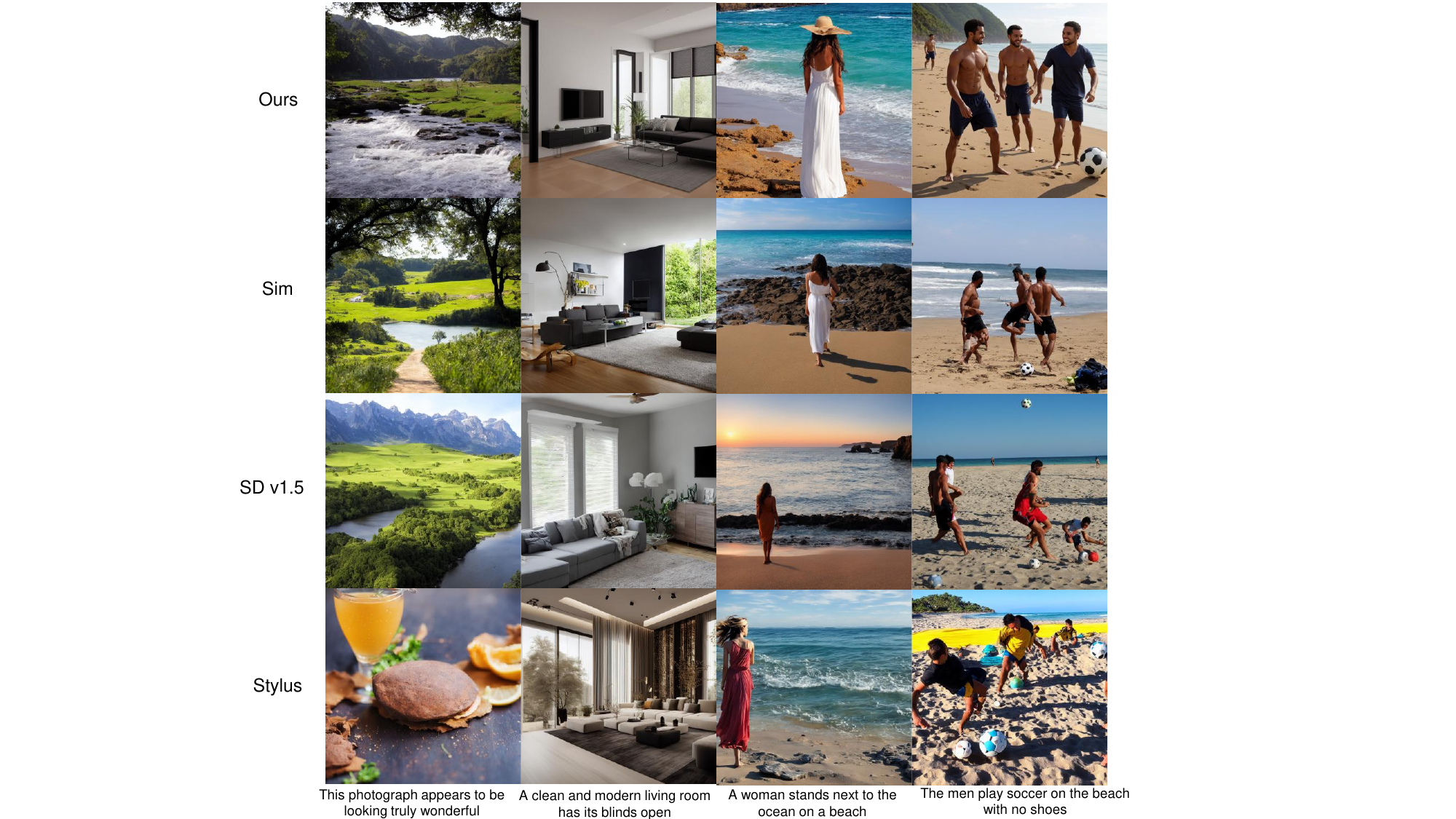}
  \caption{Images generated using our method, Euclidean distance similarity, Stable Diffusion v1.5 baseline and Stylus. Regularized linear combination exhibits comparable caption adherence and image quality.}
  \label{fig:SDpic_more}
\end{figure*}

We present the generated images in Figures \ref{fig:SDpic} and \ref{fig:SDpic_more}.
Across diverse prompts, the images produced by our method exhibits the strongest semantic alignment and visual fidelity. Our implementation builds on the Stylus codebase, which itself operates on a pruned SD v1.5 checkpoint, ensuring a fair comparison across methods. Compared to Stylus, Sim, and the base SD v1.5 model, our method consistently generates sharper and more coherent results with fewer visual artifacts. Notably, our approach better preserves object structure and context (e.g., correct limb proportions and natural scene layouts), while other methods frequently suffer from distortions or implausible object compositions. These qualitative observations support the quantitative preference scores, demonstrating that leveraging sparsity through $\ell_1$‑regularization leads to more accurate and visually appealing images.

\end{document}